\documentclass[english]{lni}

\usepackage{balance}
\usepackage{latexsym}
\usepackage{graphicx}
\graphicspath{{./images/}}
\usepackage{booktabs} % for formal tables
\usepackage{color}  % for coloring text
\usepackage{amsmath}  % for aligning equations
\usepackage{subcaption}
\usepackage{caption}
\usepackage{tikz}
\usepackage{colortbl} % for color in tables
\usepackage{framed}
\usepackage{multirow}
\usepackage{multicol}
\usepackage{url}
\usepackage{verbatim}
\usepackage{cancel}
\usepackage{xspace} % for correcting space after macro commands
\usepackage[ruled,linesnumbered,vlined]{algorithm2e}
\usepackage{bbold} % for writing mathbb{1}
\usepackage{arydshln}
\usepackage{float}

\newcommand{\struct}[1]{\texttt{\small #1}}
\newcommand{\utterance}[1]{\textit{#1}}
\newcommand{\phrase}[1]{\textit{``#1''}}

\newcommand{\myparagraph}[1]{\noindent \textbf{#1}.}

\newcommand{\squishlist}{
	\begin{list}{$\bullet$}
		{ \setlength{\itemsep}{0pt}
			\setlength{\parsep}{3pt}
			\setlength{\topsep}{3pt}
			\setlength{\partopsep}{0pt}
			\setlength{\leftmargin}{1.5em}
			\setlength{\labelwidth}{1em}
			\setlength{\labelsep}{0.5em} } }
	\newcommand{\squishend}{
\end{list}  }

\begin{document}
%%% Mehrere Autoren werden durch \and voneinander getrennt.
%%% Die Fußnote enthält die Adresse sowie eine E-Mail-Adresse.
%%% Das optionale Argument (sofern angegeben) wird für die Kopfzeile verwendet.
\title[RAGONITE]{RAGONITE: Iterative Retrieval on Induced Databases and Verbalized RDF for Conversational QA over KGs with RAG}
%% \subtitle{Untertitel / Subtitle} % if needed
 \author[1]{Rishiraj Saha Roy}{rishiraj.saha.roy@iis.fraunhofer.de}{0000-0002-5774-5658}
 \author[1]{Chris Hinze}{chris.hinze@iis.fraunhofer.de}{0009-0006-3196-6280}
 \author[1]{Joel Schlotthauer}{joel.schlotthauer@iis.fraunhofer.de}{0009-0009-6525-6094}
 \author[1]{Farzad Naderi}{farzad.naderi@iis.fraunhofer.de}{0000-0001-8226-7873}
 \author[1]{Viktor Hangya}{viktor.hangya@iis.fraunhofer.de}{0000-0002-5144-3069}
 \author[1]{Andreas Foltyn}{andreas.foltyn@iis.fraunhofer.de}{0009-0001-4678-9786}%
 \author[1]{Luzian Hahn}{luzian.hahn@iis.fraunhofer.de}{0009-0005-8450-2714}%
 \author[1]{Fabian Kuech}{fabian.kuech@iis.fraunhofer.de}{0009-0005-4665-387X}%
 % \author[2, 3]{Firstname4 Lastname4}{firstname4.lastname4@affiliation1.org}{0000-0000-0000-0000}%
 \affil[1]{Fraunhofer Institute for Integrated Circuits IIS\\Am Wolfsmantel 33\\91058 Erlangen\\Germany}
 % \affil[2]{University 2 \\Department\\Address\\Country}
 % \affil[3]{University 3\\Department\\Address\\Country}
\maketitle

\begin{abstract}
Conversational question answering (ConvQA) is a convenient means of searching over RDF knowledge graphs (KGs), where a prevalent approach is to translate natural language questions to SPARQL queries. However, SPARQL has certain shortcomings: (i) it is brittle for complex intents and conversational questions, and (ii) it is not suitable for more abstract needs. Instead, we propose a novel two-pronged system where we fuse: (i) SQL-query results over a database automatically derived from the KG, and (ii) text-search results over verbalizations of KG facts. Our pipeline supports iterative retrieval: when the results of any branch are found to be unsatisfactory, the system can automatically opt for further rounds. We put everything together in a retrieval augmented generation (RAG) setup, where an LLM generates a coherent response from accumulated search results. We demonstrate the superiority of our proposed system over several baselines on a knowledge graph of BMW automobiles.
\end{abstract}
%%% Beginn des Artikeltexts

\section{Motivation}
\label{sec:motiv}

\myparagraph{Querying knowledge graphs} It is common practice in large-scale enterprises to organize important factual data as RDF knowledge graphs (KGs, equivalently knowledge bases or KBs)~\cite{gomez2017enterprise,song2017building,galkin2017enterprise}.
Storing data as RDF KGs has the advantage of a flexible subject-predicate-object (SPO) format, which simplifies the work of moderators by eliminating the need for complex schemas as in equivalent databases.
KGs are usually queried via SPARQL, a data retrieval and manipulation language tailored to work with graph-based data. 
With the advent of large language models (LLMs)~\cite{touvron2023llama,achiam2023gpt,jiang2024mixtral,ali2024teuken}, LLMs are now used to generate SPARQL queries from the user's natural language (NL) questions~\cite{meyer2024assessing,perevalov2024towards,yang2023llm,rangel2024sparql}, replacing previous sophisticated in-house systems~\cite{bakhshi2020data,diefenbach2020qanswer,sun2020sparqa}.

\myparagraph{A two-pronged pipeline}
Like other knowledge repositories, KGs are also conveniently searched and explored via conversational question answering (ConvQA) systems~\cite{christmann2019look,saha2018complex,kaiser2021reinforcement,kacupaj2021conversational}.
However, ConvQA comes with a major challenge: vital parts of the question are left implicit by the user, as in dialogue with another human.
In preliminary experiments, we found that even with the most capable LLMs like GPT-4o,
translating a user's conversational questions to SPARQL is a bottleneck, even with representative in-context examples. This problem is exacerbated for more complex intents with conditions, comparisons and aggregations~\cite{roy2022question}. Moreover, a KG often contains information that can satisfy more abstract intents: this cannot be harnessed via SPARQL. 
In this work, we overcome these limitations and present a novel system with two branches:
\textit{SQL over induced databases} satisfy crisp information needs -- including mathematical operations,
while less defined intents are handled by \textit{text retrieval over KG verbalizations}.
Notably, we allow repeated retrievals if a single round fails to fetch satisfactory information from the backend KG.
Finally, the results of these branches are merged by a generator LLM to formulate the answer that is shown to the user who posed the conversational question. 

\section{RAGONITE: System Description and Unique Features}
\label{sec:desc}

\begin{figure}
    \centering
    \includegraphics[width=\textwidth]{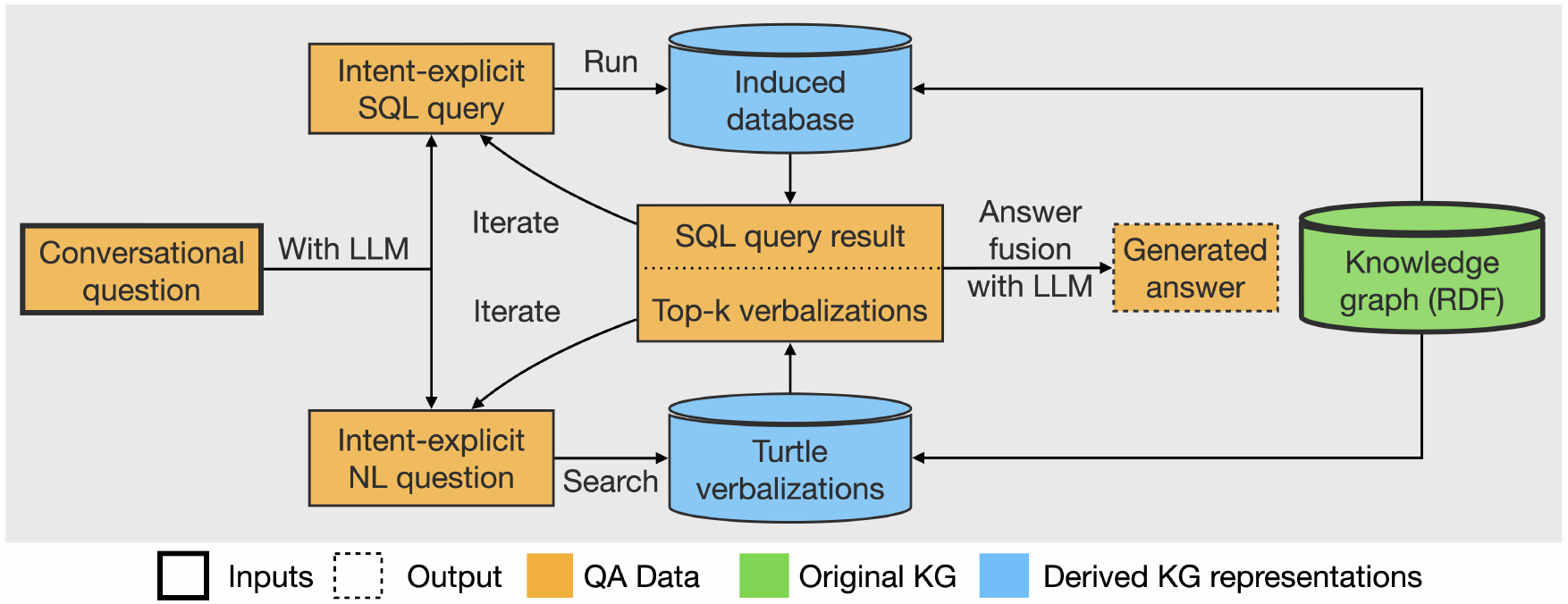}
    \caption{The RAGONITE workflow for ConvQA over KGs with retrieval augmented generation.}
    \label{fig:overview}
\end{figure}

\myparagraph{System overview} Fig.~\ref{fig:overview} shows an overview of our agentic system RAGONITE (Retrieval Augmented Generation ON ITErative retrieval results).
There are two retrieval branches in RAGONITE: (i) the SQL query executes over the DB and obtains results, and (ii) the KG verbalizations are searched via 
% lexical and
a dense retriever model~\cite{karpukhin2020dense} for passages that best satisfy the NL question.
A plausible user question to our system could be \utterance{What is the average acceleration time to 100 kmph for BMW Sport models?}, followed by another question with implicit intent, such as \utterance{And how does this compare to a typical X1 model?}
The system handles such queries by
% triggering two LLM agents in parallel,
generating an intent-explicit SQL query~\cite{li2024can} and an intent-explicit NL question~\cite{vakulenko2021question}
with an LLM.
% By an intent-explicit form, we mean a version where necessary information from the conversational context (previous questions and answers) is included in the current question to make it self-sufficient. 
An intent-explicit form is a self-sufficient variant that contains all necessary information from the previous conversational context.
The generated SQL query is then run over a database that we automatically derive from the KG (Sec.~\ref{subsec:induce}), and the NL question is searched over the verbalizations of the KG facts
% in `Turtle format'
(Sec.~\ref{subsec:verbalize}).
Comparing  SQL results and the top-k verbalizations with the conversational question and the previous history,
% a third LLM agent then
the LLM decides whether the retrieved information is sufficient to generate a satisfactory answer. If not, it suggests another round of retrieval (Sec.~\ref{subsec:iterate}). Finally, the accumulated SQL results and top-k verbalizations are fused by a second LLM agent (Sec.~\ref{subsec:fuse}) to present a coherent and fluent answer to the end user. The demo supports locally hosted LLMs as well as those accessible via APIs (Sec.~\ref{subsec:fine-tune}). %We list the unique features of our demonstration next.

% \subsection{Inducing database automatically from KG}
\subsection{Database induction}
\label{subsec:induce}

\begin{figure}
    \centering
    \includegraphics[width=\textwidth]{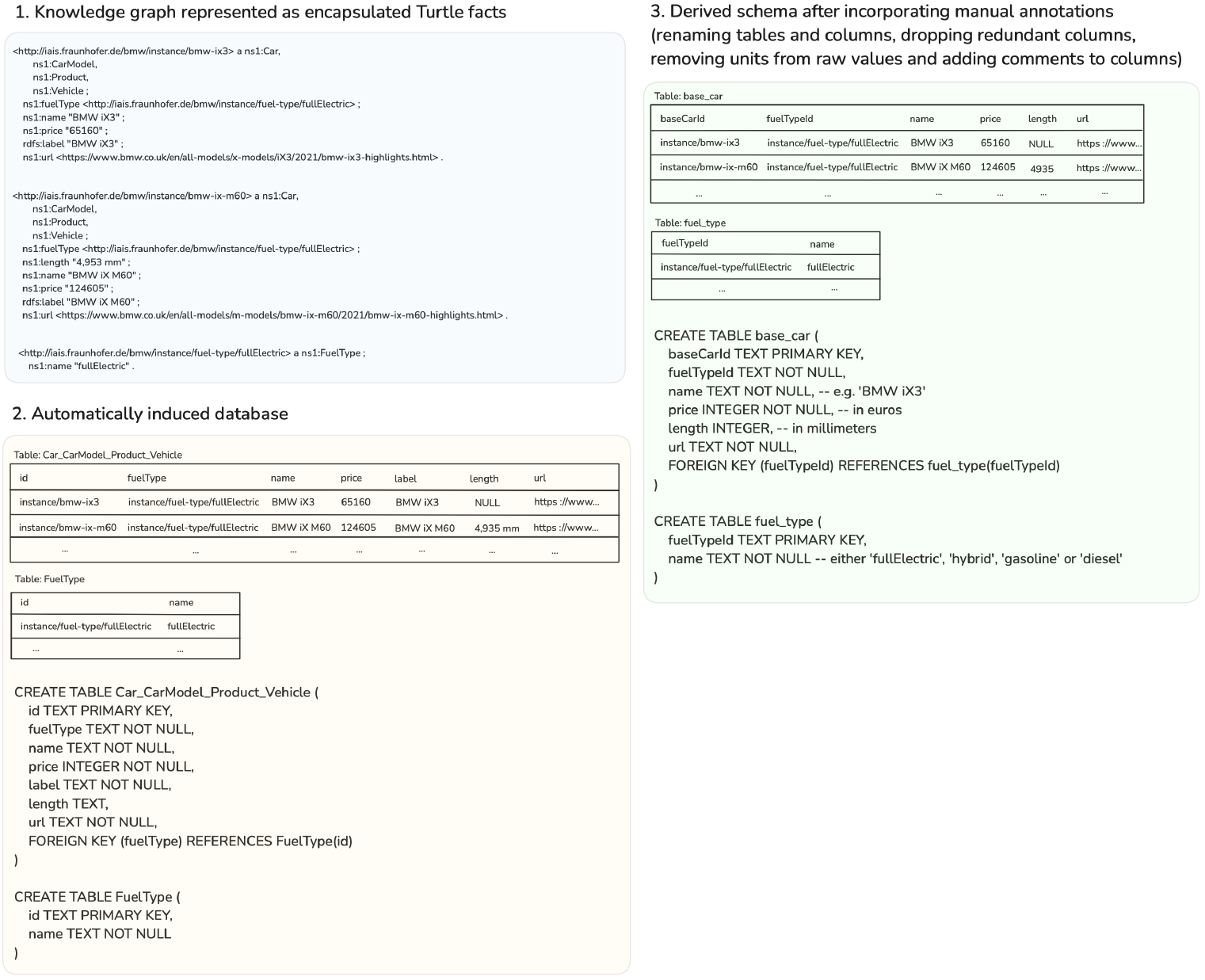}
    \caption{Canonical example illustrating automatic database induction from knowledge graph.}
    \label{fig:kg2db}
\end{figure}

A key feature of this demo is the derivation of a database from an RDF KG as follows.
% Actual KGs have long entity URIs and RDF prefixes that we do not mention here for brevity.

\myparagraph{Inducing schema}
Our KG is originally in an NTriples format\footnote{\url{https://www.w3.org/TR/n-triples/}}.
This means that it is comprised of facts that are SPO triples,
where subjects are entities (like \struct{bmw-x6-m-competition}, \struct{bmw-ix1-edrive20-sport}, or \struct{adaptive-LED-headlights}), predicates are relations (like  \struct{engine-specification}, \struct{wheelbase}, or \struct{price}), and objects are types (like \struct{car}, \struct{engine} and \struct{equipment}) or literals (constants like \struct{37450 EUR}, \struct{2760 mm} or \struct{250 kWh}).
We first convert the KG from Ntriples into Turtle format\footnote{\url{https://www.w3.org/TR/turtle/}},
that encapsulates all facts of a specific subject entity. 
We identify unique entity types based on the values of the type predicate. For each entity type a table is added to the database. The subject entity is used as the primary key and each literal value is added as a column (e.g. ``price'' or ``height'' are added to the ``car'' table). Next, the relations between entity types are analyzed. For each $1:1$ or $1:N$ relation between entity types A and B, a column is added to the entity table B that contains a foreign key which points to table A. For $N:M$ relations, an additional table is added to the database which contains two columns with foreign keys.

\myparagraph{Inserting data}
For each set of encapsulated Turtle facts, a single row is added to the respective table. The primary key is the subject entity in the Turtle facts. All literal values are added to fill the columns. Next we add the foreign keys of related entities. Then, for each table, the values of each column are analyzed to derive the data type of the column (like ``INT'' for ``price'', or ``TEXT'' for ``drive type'') and whether the values can be NULL. This information is then added to the schema.

\myparagraph{Enhancing semantics}
Finally, our system allows renaming tables and columns to improve the LLM's comprehension of the schema
(such as \phrase{engine specification} $\mapsto$ \phrase{engine}).
Additionally, 
comments can be added to each column, for example, NL explanations of complex predicate names (like \phrase{WLTPCO2Emissioncombined}).

To derive the intent-explicit SQL query from a conversational question, the LLM receives all previous questions and answers, the current question, and the database schema (all CREATE TABLE statements including data type constraints and comments). A toy example of this process is in Fig.~\ref{fig:kg2db}. Note that if one were to manually create an equivalent DB from the KG, this could be result in simpler schema: our focus in this work is on automating the process, and finding the DB with least complexity is a topic for future work.

\subsection{KG verbalization}
\label{subsec:verbalize}

Abstract questions like `\utterance{Innovative highlights in x7?}' require NL understanding and common sense reasoning to map the intent to equipments and accessories of the BMW X7. We address such questions by verbalizing KGs~\cite{oguz2022unik}, so that LLMs can reason over them like NL text.
We again use the KG in Turtle format,
where all facts with a specific entity as subject are grouped together.
The facts in each Turtle ``capsule'' are then verbalized into NL passages using simple rules:
(i) all schema-related prefixes are stripped to extract the raw content;
(ii) the subject string, predicate string and object string are concatenated to form a sentence;
(iii) \phrase{is} is added before \struct{type} predicates, and \phrase{has} before other predicates;
and (iv) The reverse formulation of the fact is also added to the passage to later help the generator LLM answer questions when the contained information is requested in complementary ways (\utterance{What is BMW 120 Sport's engine performance?} and \utterance{Which BMW engines have a performance of 125 kW?}).
Explicitly adding reversed facts is helpful for smaller LLMs.

A toy RDF Turtle capsule looks like: 
\struct{<engine/bmw-120-sport> a ns1:EngineSpecification ; ns1:enginePerformance "125 kW" ; ns1:fuelType <fuel-type/gasoline> .} This is verbalized into an NL passage as follows:
\phrase{BMW 120 Sport is Engine Specification. BMW 120 Sport has engine performance 125 kW. 125 kW is engine performance of BMW 120 Sport. BMW 120 Sport has fuel type gasoline. Gasoline is fuel type of BMW 120 Sport.}

\subsection{Iterative retrieval}
\label{subsec:iterate}

\begin{figure}
    \centering
    \includegraphics[width=\textwidth]{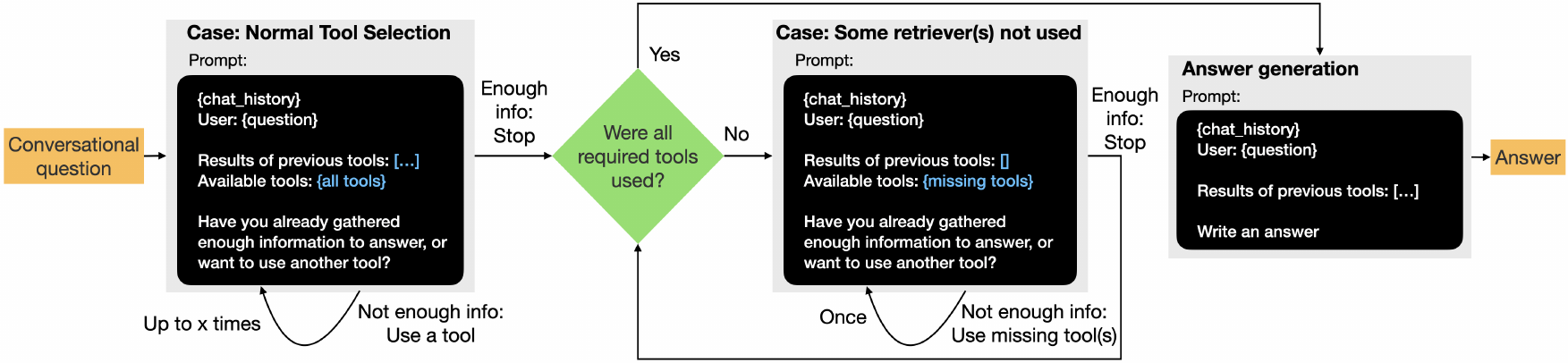}
    \caption{Illustrating iterative retrieval with SQL and text search tools via LLM calls in RAGONITE.}
    \label{fig:tool}
\end{figure}

It is possible that results from either branch of RAGONITE are not of sufficient quality. For example, the SQL tool might generate a partial query or the text results may be incomplete or contain only irrelevant information.
For such cases, the LLM can request additional rounds (set to three, but configurable) of retrieval from either branch.
Retrieval results, and possibly error messages from one round are included in the prompt in the next round. Sending such errors back to the retrievers gives them the opportunity to correct small mistakes, like
resolving an ambiguous column name (for SQL), and guiding subsequent searches towards more relevant evidence (for text search).
We further force the LLM to use 
both branches
at least once to ensure as comprehensive a response as possible.
This iterative workflow with tools is shown in Fig.~\ref{fig:tool}. 

\subsection{Branch integration}
\label{subsec:fuse}

At the end of iterative retrieval, a second LLM agent generates an answer using both the SQL outputs and the top passages being inserted into its prompt.
This way we do not have to decide which branch to prefer, neither at question-time nor after obtaining an answer from each mode: with both results at its disposal, the LLM can decide how to integrate them.
The integration of branches and judgment over individual results is thus automated via the LLM and are not hard-coded into the system.

\subsection{Open LLM support}
\label{subsec:fine-tune}

The default LLM in RAGONITE is GPT-4o with API access.
A desideratum of agentic LLM systems is to ensure data security via local, on-premise LLM hosting. Along these lines, RAGONITE supports local deployment of open LLMs such as Llama-3.3. For our experiments, we hosted the 70B version with 4 bit-quantization using ollama\footnote{\url{https://ollama.com/library/llama3.3}}.

\subsection{Heterogeneous QA}
\label{subsec:text}

Since one of our retrievers runs on NL verbalizations of the KG, it is straighforward to insert additional text contents (for example, Web documents with supplementary information to what is contained in our KG) to our backend knowledge repository. In one version of the demo, we incorporated about 400 English passages from the BMW website\footnote{\url{https://www.bmw.co.uk/en/index.html}} to increase the scope of answerable questions: RAGONITE then becomes capable of heterogeneous question answering with a mix of a KG and a text collection as its knowledge sources.

\section{Demonstration walkthrough}
\label{sec:demo}

\begin{figure}
    \centering
    \includegraphics[width=\textwidth]{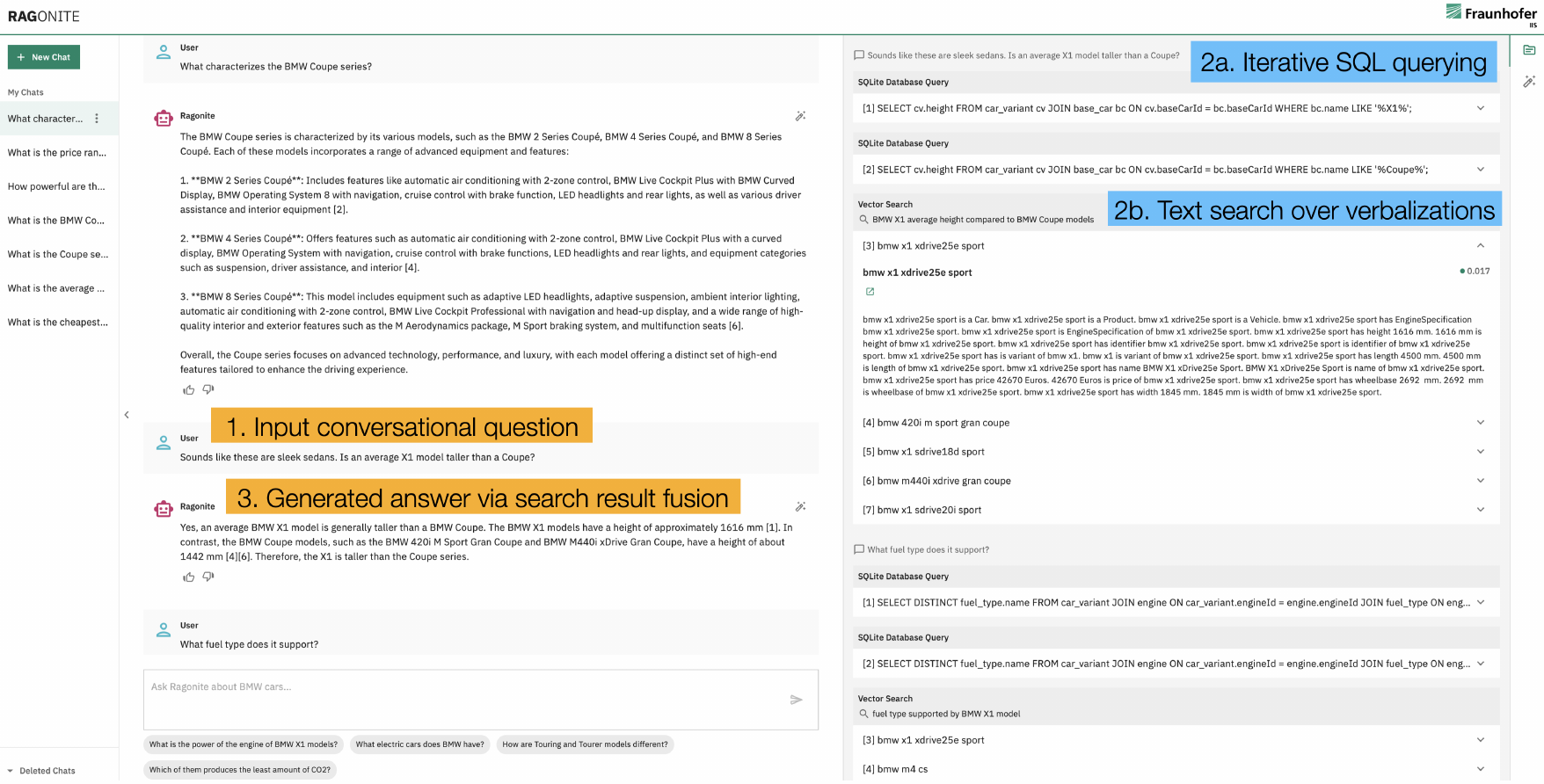}
    \caption{Screenshot of RAGONITE. Colored boxes are not part of the UI. Text readable on zoom-in.}
    \label{fig:walkthrough}
\end{figure}

\myparagraph{Overview}
% We now walk the reader through RAGONITE's user interface -- a screenshot is in Fig.~\ref{fig:walkthrough}, where orange and blue boxes are overlaid to guide the reader and are not part of the UI. We request appropriate zooming-in to read the actual text contents. 
The UI in RAGONITE (Fig.~\ref{fig:walkthrough})
% The screen
is divided vertically into four panels. Left to right, (i) the first
% narrow
panel stores previous
chats;
% conversations for future reference;
(ii) the main chat panel takes questions as input
% from the user
and displayes system answers;
% generated by the system;
(iii) the third panel shows answer derivation steps;
% helps make RAGONITE a transparent pipeline;
and (iv) the
% very narrow
last strip helps switch between different RAGONITE
configurations.
% views of the internal workings of the system and helps configure system parameters including choice of LLMs (Sec.~\ref{subsec:fine-tune}).

\myparagraph{Walkthrough} We go through only the main workflow here: everything else can be explored by a user during an interactive demo session. Offline, the backend KG is preprocessed to derive the equivalent DB and the text corpus. At runtime, in Step 1, users enter their conversational question (potentially intent-implicit) in the input box provided. Suppose we are at the second conversation turn now (\utterance{Sounds like these are sleek sedans. Is an average X1 taller than a Coupe?}). In Step 2a, this question is passed on to the intent-explicit SQL-formulating LLM call (Sec.~\ref{subsec:induce}). In this case, the LLM decides that results of one iteration (that only retrieved heights of BMW X1 variants) were not enough to satisfy the intent in this complex question (Sec.~\ref{subsec:iterate}), and so it formulates a second query (this time fetching the heights of Coupe models). At this moment, the iterator stops SQL querying and invokes the text retriever (Step 2b, Sec.~\ref{subsec:verbalize}) via an intent-explicit NL question (\utterance{BMW X1 average height compared to BMW Coupe Models}). Note that Steps 2a and 2b can happen in an arbitrary order. A sample verbalization of a Turtle capsule can be seen in the screenshot below the corresponding blue box (for the entity \texttt{bmw-x1-xdrive-25e-sport}, we lowercase all text for simplicity), along with the retrieval score assigned by the vector search model ($0.017$). The iterator decides that contents of one round of text retrieval has enough information pertinent to the question, and by now both tools have been explored, so it hands over these results to the answer generator (Sec.~\ref{subsec:fuse})  This LLM agent crafts the final response shown to the user (Step 3). Through explicit prompting, we make the LLM inject source citations in its answer for scrutability. In this case, sources [1-2] refer to SQL results and [3-7] to verbalizations (understandable by labels above respective results).
Prompts for all LLM calls can be examined by an user via the answer derivation panel.

\myparagraph{Backend}
The core of RAGonite's backend handles query formulation, SQL execution, text retrieval, iteration/tool selection, and answer generation. The backend also has a stateful layer that stores conversations in a SQLite DB and provides a REST API to the frontend with FastAPI.
Core dependencies include the vector database (ChromaDB), a template engine used for prompts (Jinja), and LLM libraries (OpenAI and Ollama).
% Standard Python libraries were used for SQL.
ChromaDB is used as our vector DB for storing RDF verbalizations. We also use ChromaDB's in-built vector search.
% While ChromaDB was used for its extensibility, we also explored variants like Weaviate and Milvus.
% A single GPU server (4x48GB NVIDIA Ada 6000 RTX, 512 GB RAM, 64 virtual cores) was used for all our experiments. 
For all of our experiments, we ran RAGONITE on a shared GPU server (4x48GB NVIDIA Ada 6000 RTX, 512 GB RAM, 64 virtual cores).
% which is shared with other researchers.
A single GPU was dedicated to running Llama-3.3-70B. If the LLM used for text inference is hosted in the cloud, for example, by using an OpenAI model, the remaining resource requirements are minimal and the RAGonite instance can also run on a standard laptop.

\myparagraph{Frontend}
We created a single-page React app, intentionally avoiding further dependencies to preclude the need for a build process. All API calls are handled by the frontend.
\section{Evaluation}
\label{sec:res}

\myparagraph{Setup} Our BMW KG (scraped from \texttt{bmw.co.uk}) has $3442$ facts, $466$ unique entities, $27$ predicates, $7$ types, and $1295$ literals. So we get $7$ tables and $466$ passages via DB induction and RDF verbalization. The authors generated $6$ conversations with $5$ turns each, and test $4$ GPT-4o-based configurations with them: (i) SPARQL-only, (ii) SQL-only, (iii) verbalization-only, and (iv) SQL+verbalizations. Questions were in $3$ categories ($10$ in each): (i) lookup intents (\utterance{cost of 530em sport saloon?}), (ii) complex intents (\utterance{charging time of 225e Active Tourer less than avg over all bmws?}), and (iii) abstract intents (\utterance{luxury features in x5?}). Answers were marked correct/incorrect (SQL/SPARQL query must also be correct). 

\myparagraph{Results} The two-pronged iterative approach did the best ($28/30$ questions correct).
SQL-only got $18$ (failure to handle abstract intents) and verbalization-only got $24$ cases right (failure to handle complex math). A baseline where GPT-4o generated SPARQL queries, got only $4$ answers correct. 
A particular vulnerability of SQL and SPARQL was incorrect entity linking for ad hoc mentions (like \phrase{gran coupe sport 220 i m} in questions, while the exact entity label is \struct{bmw-220i-m-sport-gran-coup\'{e}}), and this is where verbalizations are particularly helpful.
This also proved that LLMs are still much weaker at SPARQL generation than SQL (presumably due to lower training data), especially when it comes to complex or abstract intents in conversational question answering.

\myparagraph{Runtimes} We recorded stepwise runtimes for each of our $30$ benchmark questions using our default pipeline with GPT-4o. 
Including question completion, the entire retrieval took $3.936$ seconds, and answering took $2.409$ seconds.
Zooming in on the retrieval step, tool selection took $1.047$ seconds, SQL execution took $0.095$ seconds, and vector search took $0.539$ seconds.
The entire RAG pipeline thus took $6.348$ seconds per question, on average. RAGONITE thus operates with interactive response times.

All public artifacts (like code, data, and demo screenshots) related to this work can be found at \texttt{\url{https://github.com/Fraunhofer-IIS/RAGonite}}.

\section{Conclusions and future work}
\label{sec:confut}

We showcase RAGONITE, a transparent RAG pipeline that adopts a novel alternative approach to ConvQA over KGs by iterative retrieval and merging results of structured and unstructured querying. 
While Text2SPARQL has been investigated in the past, we adopt a novel alternative of exploiting Text2SQL, where LLMs do much better, over a DB that is programmatically induced from the KG.
By converting a KG to its verbalized form that is indexed as a vector DB, we show that the information in knowledge graphs can be useful for more abstract intents as well. RAGONITE contains an agentic workflow: key future work would enhance this further by incorporating reflection mechanisms that enable each module to critique and improve its output, without compromising efficiency. Another direction would be to fine-tune specific pipeline components for a RAG setting with synthetic data.

\section*{Acknowledgements}

% We would like to acknowledge the Fraunhofer OpenGPT-X team for their support. 
This research was funded by the German Federal Ministry for Economic Affairs and Climate Action (BMWK) through the project OpenGPT-X (project no. 68GX21007D).
We thank Lucas Weber from the NLP team at Fraunhofer IIS for useful comments on an initial version of this manuscript. 
We also thank the rest of the team members for useful inputs at various stages of this work, intensively testing our demo, and suggesting benchmark questions.
% We acknowledge Afshin Sadeghi, Sangamithra Panneer Selvam, and other members from Fraunhofer IAIS for help with an early version of the KG taxonomy.

%%% Angabe der .bib-Datei (ohne Endung) / State .bib file (im Falle der Nutzung von BibTeX)
\bibliography{2025-arxiv-dp-ragonite-bmw}
%% \printbibliography % im Falle der Nutzung von biblatex
\end{document}